\def\year{2019}\relax
\documentclass[letterpaper]{article}
\usepackage{aaai19}
\usepackage{times}
\usepackage{helvet}
\usepackage{courier}
\usepackage{url}
\usepackage{graphicx}
\usepackage{amsmath}
\usepackage{algorithm2e}
\usepackage[table]{xcolor}

\title{Improving Search with Supervised Learning in Trick-Based Card Games}
\author{
Christopher Solinas \and Douglas Rebstock \and Michael Buro\\
Department of Computing Science, University of Alberta\\ 
Edmonton, Canada\\
\{solinas,drebstoc,mburo\}@ualberta.ca
}

\begin{document}
\maketitle
\begin{abstract}
In trick-taking card games, a two-step process of state sampling and evaluation 
is widely used to approximate move values. 
While the evaluation component is vital, the accuracy of move value 
estimates is also fundamentally linked to how well the sampling distribution 
corresponds the true distribution. 
Despite this, recent work in trick-taking card game AI has mainly focused on 
improving evaluation algorithms with limited work on improving
sampling. In this paper, we focus on the effect of sampling on the 
strength of a player and propose a novel method of sampling more 
realistic states given move history. 
In particular, we use predictions about locations of individual cards made by a
deep neural network --- trained on data from human gameplay --- in order to
sample likely worlds for evaluation. This technique, used in conjunction
with Perfect Information Monte Carlo (PIMC) search, provides a substantial
increase in cardplay strength in the popular trick-taking card game of Skat.
\end{abstract}


\section{Introduction} \label{sec:intro}

Games have always been at the forefront of AI research 
because they provide a controlled, efficient, and predictable environment to 
study decision-making in various contexts. 
Researchers are constantly trying to develop new techniques that allow for 
intelligent decisions to be made in more complex environments. 

Imperfect information games require players to make decisions without being 
able to observe the full state. 
This setting is attractive to researchers because it is a closer approximation
to real life.
To make good decisions, the player to move must account for the private 
holdings of  every player in the game, and must consider that each action they 
take reveals information about their holdings.
These are just some of the reasons that algorithms developed for perfect 
information games may not translate well to games with hidden information.

For imperfect information games with information sets too large to game
theoretically solve, most game-playing algorithms can be broken down into two
key components: inference and state evaluation. 
State evaluation tells a player how advantageous a particular state is; whereas
inference allows a player to determine the likelihood of said state. 
Both components are vital to good play. 
Even if a player could perfectly evaluate every state in an information set, 
combining them with inaccurate inference can easily introduce catastrophic 
errors in the final move values. 
In trick-taking card games like Contract Bridge or Skat, private information 
slowly becomes public as cards are played, but every move each player makes can
reveal additional information through careful reasoning. 
The most experienced human players are able to read into the implications 
of every opponent move and act accordingly.

In this paper, we investigate inference in trick-taking card games.
We show that the history of moves in the game thus far can be used to infer the
hidden cards of other players, and that this information can be used to 
considerably boost the performance of simple search-based evaluation techniques
in the domain of Skat.

The rest of this paper is organized as follows.
We first summarize the state of research related to state evaluation and 
inference for cardplay in trick-taking card games, and explain the basic rules 
of Skat --- our application domain.
Next, we describe a technique for performing state inference by predicting the 
locations of individual cards using move history. 
Then, we provide statistically significant experimental results that show the
effectiveness of this type of inference in trick-based card games like Skat.
Finally, we finish the paper with conclusions and ideas for future research.


\section{Background and Related Work} \label{sec:bg}

Card games have long been an important application for researchers wishing to
study imperfect information games. 
Recent advances in computer poker 
\cite{moravvcik2017deepstack,brown2017superhuman} have led to 
theoretically sound agents which are able to outperform professional human 
players in the full variant of Poker called Heads Up No Limit Texas Hold'em.  
However, the same techniques have thus far been unable to find success in 
trick-taking card games because they rely on expert abstractions in order
to scale to such large games.
Compact yet expressive abstractions are difficult to construct in trick-taking
card games because every single card in each player's hand can immensely affect
the value of a state or action.

Perfect Information Monte Carlo (PIMC) search \cite{levy1989million} has been 
successfully applied to popular trick-taking card games like Contract 
Bridge \cite{ginsberg2001gib}, Skat \cite{buro2009improving}, Hearts and 
Spades \cite{sturtevant2008analysis}.
PIMC has been heavily criticized over the years, starting with Frank and Basin 
\shortcite{frank1998search} because it naively evades the imperfect information 
elements of the game tree.
However, it has remained relevant because it is still among the 
state-of-the-art algorithms for these games.
In Long et al. \shortcite{long2010understanding} the authors attempt to
understand this success and conclude that for classes of games with certain
properties, including trick-taking card games, 
``PIMC will not suffer large losses in comparison to a game-theoretic 
solution.''

Furtak and Buro \shortcite{furtak2013recursive} implement a recursive
variant of PIMC (IIMC) to alleviate some of the issues pointed out by
Frank and Basin --- resulting in the current state-of-the-art player for Skat.
Elsewhere, Information Set Monte Carlo Tree Search (ISMCTS) 
\cite{cowling2012information} addresses the
same issues in three different domains, but as Furtak and Buro 
\shortcite{furtak2013recursive} argue, the resulting move values are biased 
because the player leaks private information to their playout adversaries by 
only sampling states consistent  with the player's private information 
and allowing the strategies of the playout adversaries to adapt across rollouts.
Sampling inconsistent states makes the search space intractable for many 
applications.
Baier et al. \shortcite{baier2018emulating} proposes a method for biasing 
MCTS techniques by boosting the scores of nodes reached by following actions
that are judged likely to be played by humans according to a supervised model.
Applying this to ISMCTS in the imperfect information setting is 
straightforward, but the resulting algorithm neglects the action history
occurring before the root of the search and samples states uniformly from root
information set before proceeding.
Each of these contributions improve state evaluation quality, but they fail to 
address the sampling problem investigated in this work.

Kermit \cite{buro2009improving,furtak2013recursive} uses a table-based
procedure that takes opponent bids or declarations into account in order to
infer the likelihood of states within an information set. 
Unlike our work, Kermit does not use the sequence of actions during the 
cardplay phase for further inference --- only marginalizing over its own 
private cards and those that have already been played.
Ginsberg's bridge-playing GIB \cite{ginsberg2001gib} was the first 
successful application of PIMC in a trick-taking card game.
GIB also appears to perform some state inference in that it samples ``a set $D$
of deals consistent with both the bidding and play'', but details regarding the
inference are absent from the paper.

In other domains, Richards and Amir \shortcite{richards2007opponent} used
opponents' previous moves to infer their remaining tiles in Scrabble.
Their program samples and weighs possible sets of letters that could remain
on an opponent's rack after a move --- assuming that the opponent made the 
highest ranked play according to their static evaluation.
  
\subsection{Skat}

Though particularly popular in Germany, Skat is a 3-player card game that is
played competitively in clubs around the world.
Each player is dealt 10 cards from a 32-card deck, and the remaining two 
(called the skat) are dealt face down. 
Players earn points by winning rounds which can be broken down into two main 
phases: bidding and cardplay.

In the bidding phase, players make successively higher bids to see who will
become the soloist for the round. 
Playing as the soloist means playing against the other two players during 
cardplay and carrying the risk of losing double the amount of possible points 
gained by a win. 
The soloist has the advantage of being able to pick up the skat and discard two 
of their 12 cards.
The soloist then declares which of the possible game types will be played for
the round. 
Standard rules include suit games (where the 4 jacks and a suit chosen by the 
soloist form the trump suit), grands (where only the 4 jacks are trump), and
nulls (where there are no trump and the soloist must lose every trick to win). 
In suit and grand games, players get points for winning tricks containing 
certain cards during the cardplay phase. 
Unless otherwise stated, the soloist must get 61 out of the possible 120 card 
points in order to win suit or grand games.
The number of points gained or lost by the soloist depends the game's base 
value and a variety of multipliers.
Most of the multipliers are gained by the soloist having certain configurations
of jacks in their hand.
The game value (base $\times$ multiplier) is also the highest possible bid the
soloist can have made without automatically losing the game.

Cardplay consists of 10 tricks in which the trick leader (either the player
who won the previous trick or the player to the left of the dealer in the
first trick) plays the first card. 
Play continues clockwise around the table until each player has played. 
Players may not pass and must play a card of the same suit as the leader if 
they can --- otherwise any card can be played. 
The winner of the trick is the player who played the highest card in the led 
suit or the highest trump card. 
Play continues until there are no cards remaining, and then the outcome of the 
game is decided. 
Many of the details of this complex game have been omitted because they are not 
required to help understand this work. 
For more in-depth explanation about the rules of Skat we refer interested 
readers to {\small\url{https://www.pagat.com/schafk/skat.html}}.

One of the main challenges with developing search-based algorithms that can
play Skat at a level on par with human experts is the number and size of the
information sets within the game. 
For instance if a player is leading the first trick as a defender, there are 
over 42 million possible states in the player's information set. 
Other challenges include playing cooperatively with the other defender when on 
defense in order to have the best chance of beating the soloist. 
This requires players to infer opponent's and partner's cards, and human 
experts resort to intricate signalling patterns to pass information to their 
teammates.


\section{Guiding PIMC search with Cardplay Inference} \label{sec:inf}

This paper is built on the foundation that performing inference on move 
history is not only possible in these types of games, but also useful.
In this section we propose a technique for state inference in imperfect
information games, and demonstrate how to apply it to improve play.

\subsection{PIMC+ Search}

Algorithm \ref{alg:pimc_plus} shows the basic PIMC algorithm, modified so
that evaluated states are sampled from an estimated distribution based on move 
history $h$.
Resulting move values are averaged over all evaluated states, so improving the 
state probability estimate has the potential to increase PIMC's playing 
strength considerably.

In particularly large information sets, estimating probabilities for the
entire set of states may become intractable. 
However, because this algorithm is easily parallelizable, in practice, 
information sets must contain billions of states before this becomes a problem 
on modern hardware.
In these cases, uniformly sampling a subset of states without replacement as an
approximation for the complete information set should be sufficient.

\begin{algorithm}[t]
  \DontPrintSemicolon
  \SetKwFunction{FMain}{PIMC}
  \SetKwFunction{sample}{Sample}{}
  \SetKwFunction{moves}{Moves}{}
  \SetKwFunction{weight}{ProbabilityEstimate}{}
  \SetKwFunction{norm}{Normalize}{}
  \SetKwFunction{perf}{PerfectInfoVal}{}
  \SetKwProg{Fn}{Algorithm}{}{}
  \Fn{\FMain{\textrm{InfoSet} $I$, \textrm{int} $n$, \textrm{History} $h$}}
  {
    \For{$m \in \moves{$I$}$} {
      $v[m] = 0$
    }
    \For{$s \in I$} {
      $p[s] \leftarrow \weight{$s, h$}$
    }
    $p \leftarrow \norm{$p$}$ \;
    \For{$i \in \{1..n\}$}
    {
      $s \leftarrow \sample{$I,p$}$ \;
      \For{$m \in \moves{$I$}$}
      {
        $v[m] \leftarrow v[m] + \perf{$s, m$}$
      }
    }
  }
\caption{PIMC with state inference}
\label{alg:pimc_plus}
\end{algorithm}

\begin{figure*}
  \centering
  \includegraphics[width=0.8\textwidth]{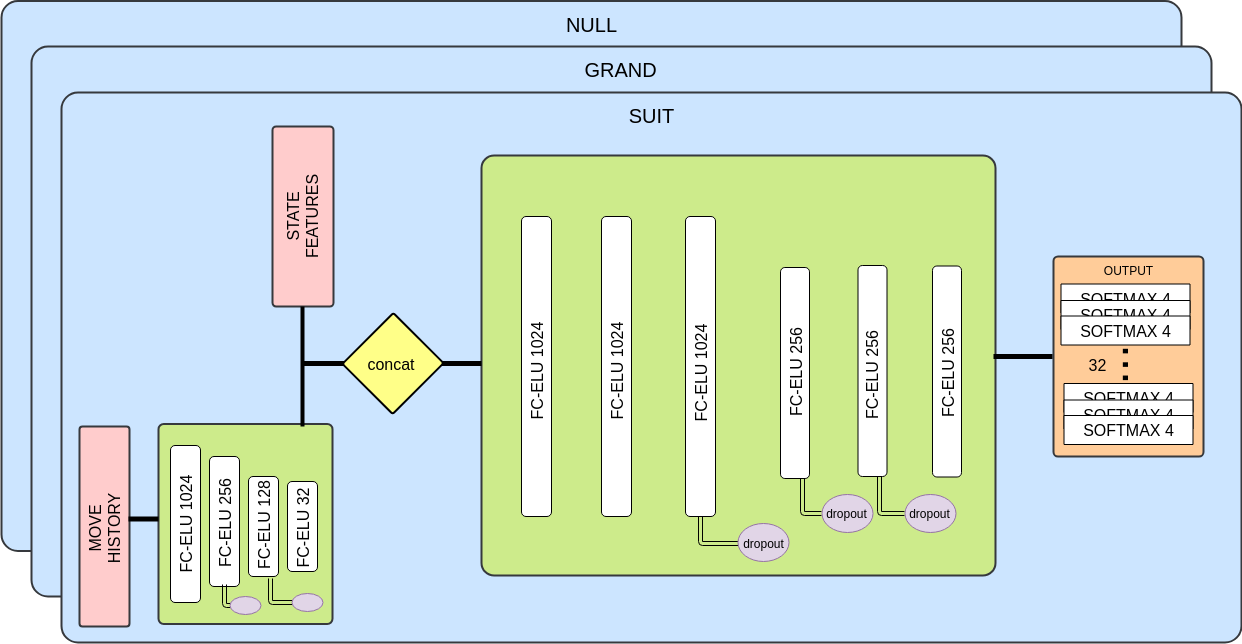}
  \caption{Inference network architecture. Shown is a Skat-specific
    architecture for predicting the pre-cardplay locations of all 32
    individual cards. Each card can be in one of four possible locations 
    (each player's hand and the skat).
    Output targets have 10 cards in each player's hand and
    the remaining 2 in the skat.}
  \label{fig:network}
\end{figure*}

\subsection{Individual Card Inference with Neural Networks}

Previous work in trick-taking card games uses table-based approaches for 
inference. 
This works well if the context is small enough that there is sufficient data 
corresponding to every table entry.  
However, as the context grows larger and the amount of training data for each 
context declines, table-based approaches become more prone to overfitting and 
more difficult to work with.
Eventually it becomes necessary to generalize across contexts in order to 
make good predictions.

Neural networks are well suited for this type of problem, but in our case 
the straightforward approach of predicting the state directly is too difficult
because of the sheer number of possible states to consider.
In order to make the problem tractable, we propose a method that instead
predicts the locations of individual cards.

To turn predictions about individual card locations into a probability
distribution of the possible states in an information set, we apply
Equation~\ref{eq:world_prob_single}. 
Assuming independence, we multiply the probabilities of each card $c$'s true 
location in state $s$ given move history $h$. 
This provides a measure for each state that can be normalized into a 
probability distribution for the set of states $S$ in the information set:

\begin{equation} \label{eq:world_prob_single}
  p(s|h) \propto \prod_{c\in C} L(h)_{c,loc(c,s)}
\end{equation}

\noindent
where $C$ is the set of cards, $L(h)$ is a real $|C| \times l$ matrix, $l$ is
the number of possible card locations, and $loc(c,s)$
is the location of card $c$ in state $s$. 
In Skat,  $|C|=32$ , $l=4$, and $loc(c,s) \in \{0,1,2,3\}$  because there are 32
cards and each card is located either in hand 0, 1, 2, or the skat.
Entries of $L(h)$ are estimated probabilities of cards in locations given move
history $h$.

In this framework it is possible to provide structure to the network
output in order to capture game-specific elements. 
For instance, taking softmax over the rows of $L$ constrains the probability 
masses so that the  sum for each card adds up to 1. 
Our work does not impose any additional constraints, but constraints on the 
number of total cards in each hand or each suit's length could be added as well. 

One practical insight for making this type of prediction is that learning 
is easier when the full targets are used instead of just predicting 
the unknown elements.
In our case, this means predicting the full 32-card configuration rather than
only trying to predict the missing cards.

The next section describes how we approach feature engineering and network 
design in Skat, but it can be adapted to other games using domain-specific 
knowledge.
It is important to consider how to incorporate the move history and other game 
state features to allow the network to learn a good representation. 
Sequences in Skat are relatively short, so we are able to have success using a 
simple fully-connected network. 
However, our approach is not dependent on such details and more complex 
constructs, such as recurrent units to capture dependencies in longer sequences, 
should be considered if they fit the problem.

\subsection{Application to Skat}

Figure~\ref{fig:network} details our network architecture.
We train a separate network for each game type (suit, grand, and null). 
Regardless of the game type, there are 32 total cards in Skat that can be in 
any of 4 potential positions (3 hands and the skat), 
Each network has the same overall structure. 
We use dropout \cite{srivastava2014dropout} of 0.8 on layers 2, 3, and 4 and 
early-stopping \cite{prechelt1998automatic} on a validation set to reduce 
overfitting.  
Table~\ref{tab:hyperparams} lists all hyperparameters used during training.
Hidden layers use ELU activations \cite{clevert2015fast}, and the network is 
trained by minimizing the average cross-entropy of each card output.

\begin{table}[b]
  \centering
  \caption{Network training hyper-parameters}
  \label{tab:hyperparams}
  \begin{tabular}{c|c}
    Parameter & Value \\
    \hline
    Dropout & 0.8 \\
    Batch Size & 32 \\
    Optimizer & ADAM \\
    Learning Rate (LR) & $10^{-4}$ \\
    LR Exponential Decay& 0.96 / 10,000,000 batches \\
  \end{tabular}
\end{table}

\begin{table}[b]
  \centering
  \caption{Network input features}
  \label{tab:features}
  \begin{tabular}{c c}
    Feature & Width \\
    \hline
    Player Hand & 32 \\
    Skat & 32 \\
    Played Cards (Player, Opponent 1\&2) & 32*3 \\
    Lead Cards (Opponent 1\&2) & 32*2 \\
    Sloughed Cards (Opponent 1\&2) & 32*2\\
    Void Suits (Opponent 1\&2) & 5*2 \\
    Max Bid Type (Opponent 1\&2) & 6*2 \\
    Max Bid Magnitude (Opponent 1\&2) & 5*2 \\
    Current Trick & 32 \\
    Soloist & 3 \\
    Trump Suit & 5 \\
    Cardplay History & 32*24\\
  \end{tabular}
\end{table}

We use various input features to represent the state of the game in the view
of the player to move --- they are listed in Table~\ref{tab:features}.  Lead
cards are the first cards played in a trick, and sloughed cards are those that
are played when a player cannot follow suit but also does not a trump card. 
Void suits indicate when players' actions have shown they cannot have a suit
according to the rules of the game.
Bidding features are broken down into type and magnitude.  
Type indicates a guess as to which game type the opponent 
intended to play had they won the bidding with their highest bid.
This is computed by checking if the bidding value is a multiple of the 
game type base value.
Magnitude buckets the bidding value into 1 of 5 ranges that are intended to 
capture which hand multiplier the opponent possesses. 
Domain knowledge is used to construct ranges that group different base game
values with the same multiplier together.
The exact ranges used are $18..24, 27..36, 40..48, 50..72, \ \textrm{and} >72$.
These ranges contain some unavoidable ambiguity because some bids are divisible
by multiple game values, but bid multiplier is a strong predictor for the 
locations of jacks in particular.  
The soloist and trump suit features indicate which player is the soloist and 
which suit is trump for the current game, respectively.
All of the above features are one-hot encoded.

Due to its length, we provide the entire cardplay history (padded with zeros for
future moves) as a separate input to the network.  
This input is fed through 4 separate hidden layers that reduce its 
dimensionality to 32, at which point it is concatenated with the rest of the 
state input features and fed through the rest of the network.

The networks are trained using a total of 20 million games played by humans on a 
popular Skat server \cite{doskv2018skat}.
A design decision was made to only train the network to make predictions about 
the  first 8 tricks because information gained by inference in the last tricks 
is minimal beyond what is already known by considering void suits.
The entire training process uses Python Tensorflow \cite{abadi2016tensorflow}.

The network output is used as described in Equation~\ref{eq:world_prob_single} 
to compute probabilities for Algorithm~\ref{alg:pimc_plus}, and 
likely states are sampled for evaluation with PIMC.
As previously mentioned, the algorithm is computationally expensive
in early tricks where the information sets are relatively large.  
The rough maximum of 42 million states in Skat is manageable in around 2
seconds on modern hardware and our current implementation could be 
parallelized further. 
It should be noted that this process only performs a single forward pass of the
network per information set, so the performance bottleneck is in 
multiplying the card probabilities for each state and normalizing the 
distribution.



\begin{figure*}
  \centering
  \includegraphics[width=0.8\textwidth]{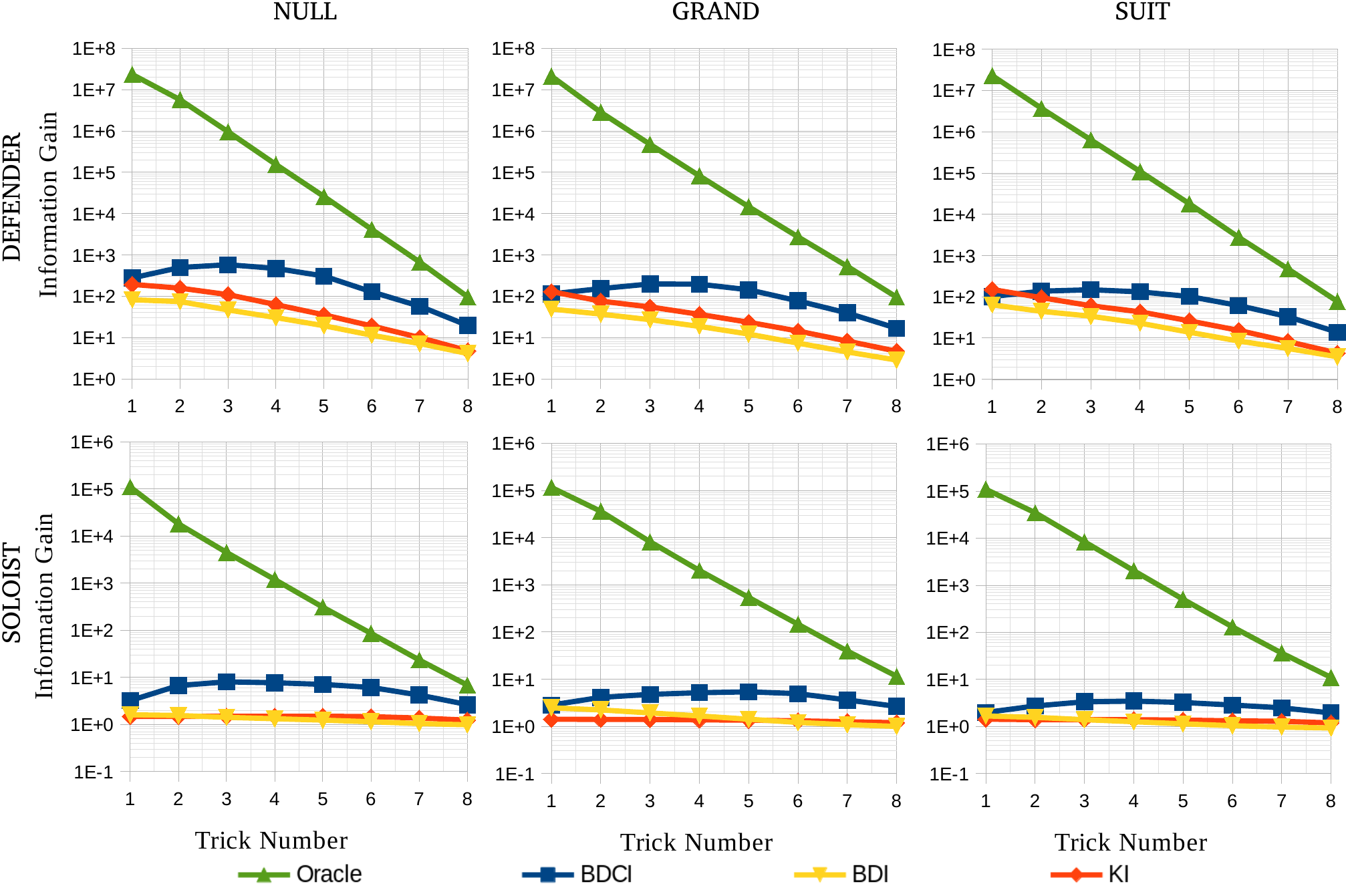}
  \caption{Average \textit{TSSR} for BDCI, BDI, KI, and Oracle inference, separated by 
  trick number for each game type (null, grand, suit) and role 
  (defender, soloist) combination. }
  \label{fig:infResults}
\end{figure*}

\section{Experiments} \label{sec:exp}

We use two methods of measuring inference performance in this work. 
First, we measure the quality of our inference technique in isolation using 
a novel metric. 
Second, we show the effect of using inference in a card player by running 
tournaments against several baseline players.

All baseline players use PIMC for evaluation and only vary in how they select 
states to evaluate. \textbf{BDCI} (``Bidding-Declaration-Cardplay Inference'') 
uses our method of individual-card inference to build a distribution from which 
states are sampled based on bidding and cardplay.
\textbf{NI} (``No Inference'') samples states uniformly from the information 
set. 
This player uses no inference of any kind to guide sampling, so it should be 
considered as a true baseline for PIMC. 
\textbf{KI} (``Kermit's Inference'') is Kermit's SD version (``Soloist/defender
inference'') described in \cite{buro2009improving} which is considered the
state-of-the-art for PIMC-based Skat players. 
\textbf{BDI} (``Bidding-Declaration Inference'') is a baseline that performs 
the same process as BDCI, but only considers bidding and game declaration
information to predict the locations of individual cards. 
This player was created to control for any possible effects of our new sampling 
algorithm (no previous work considers computing probabilities for all 
states in the larger information sets). 
We expect this player to perform comparably to \text{KI}.

\subsection{Inference Performance}

In order to directly measure inference performance, we
compare the probability of sampling the true state from the computed
distribution and the probability of uniformly sampling it from the information
set. 
This comparison provides the True State Sampling Ratio (\textit{TSSR}) which conveys 
how many times more likely the true state is going to be selected, 
compared to uniform random sampling.
\begin{equation} \label{eq:tssr}
  \mathit{TSSR} ~=~ p(s^*|h)~/~(1/n) ~=~ p(s^*|h) \cdot n
\end{equation}
\noindent
$p(s^*|h)$ is the probability the true state is sampled given the
history, and $n$ is the number of possible states. For BDI and BDCI, this is 
calculated directly.
For KI, the expectation of $p(s^*|h)$ is determined using a 
Monte-Carlo estimate.

To evaluate inference performance in isolation, \textit{TSSR} is calculated for each 
algorithm and each trick, with defender and soloist and game-types separated. 
Each trick number (1 through 8), role (defender or soloist), and game type were 
evaluated for each algorithm on 3,000 samples from holdout sets containing
games previously played by humans.

Figure \ref{fig:infResults} shows the average value of \textit{TSSR} for the 
algorithms  as well as a strict upper bound for \textit{TSSR} dubbed the 
\textbf{Oracle}. 
The Oracle predicts the true world's probability to be 1.0, so its 
\textit{TSSR} value is equivalent to the total number of possible worlds. 
The value of \textit{TSSR} is markedly impacted by both game type and player role.
For all algorithms, \textit{TSSR} is uniformly larger for defender compared to soloist. 
This is due to the declarer choosing the game that fits their cards, making 
inference much easier for the defender. 
Furthermore, the soloist knows the locations of more cards to begin with 
because they know the skat --- meaning that there is less potential for 
inference in the first place.
 
For BDI and KI, the average value of \textit{TSSR} reaches its peak in early tricks
and decreases over time. 
BDCI, however, peaks around tricks 3-5, and performs consistently better in 
the mid-game.
We attribute this to the inclusion of move history in prediction.
With the exception of the very first trick for grand and suit defender games, 
BDCI  performs considerably better in terms of \textit{TSSR}. 
Due to their reliance on the same input features, the baselines of 
KI and BDI perform comparably as expected.  

\begin{table*}[t]
  \centering
  \resizebox{\textwidth}{!}{%
  \begin{tabular}{|c||c|c|c|c|c|c|c|c|c|}
    \hline
    Game Type& \multicolumn{3}{c|}{Suit} & \multicolumn{3}{c|}{Grand} & \multicolumn{3}{c|}{Null}\\\hline
    \#Samples & 80 & 160 & 320 & 80 & 160 & 320 & 80 & 160 & 320\\\hline\hline
    BDI : NI & 22.8 : 19.0 & 21.9 : 19.4 & 22.5 : 18.9 & 39.1 : 37.7 & 38.3 : 37.9 & 38.5 : 38.1 & 15.9 : 11.8 & 14.9 : 13.2 & 16.1 : 11.7\\
    $\Delta$ & 3.8 & 2.5 & 3.6 & 1.4 & 0.4 & 0.5 & 4.1 & 1.7 & 4.4\\\hline
    KI : NI & 23.9 : 17.9 & 22.8 : 18.4 & 22.7 : 18.1 & 38.8 : 37.3 & 38.6 : 37.3 & 38.5 : 36.6 & 16.9 : 10.7 & 16.2 : 11.0 & 16.6 : 10.8\\
    $\Delta$ & 6.0 & 4.4 & 4.6 & 1.6 & 1.3 & 1.9 & 6.2 & 5.2 & 5.8\\\hline
    BDCI : NI & 25.0 : 15.6 & 23.8 : 16.3 & 24.0 : 15.7 & 39.2 : 35.5 & 39.2 : 35.4 & 39.0 : 35.0 & 17.2 : 8.3 & 17.9 : 7.9 & 18.2 : 7.5\\
    $\Delta$ & 9.4 & 7.6 & 8.3 & 3.7 & 3.7 & 4.0 & 8.9 & 9.9 & 10.8\\\hline
    KI : BDI & 21.2 : 19.2 & 20.0 : 19.5 & 20.1 : 19.7 & 37.8 : 37.5 & 38.2 : 36.8 & 37.8 : 37.3 & 13.6 : 12.4 & 13.7 : 12.4 & 13.3 : 12.1\\
    $\Delta$ & 2.1 & 0.5 & 0.4 & 0.3 & 1.4 & 0.5 & 1.1 & 1.3 & 1.3\\\hline
    BDCI : BDI & 21.9 : 16.9 & 21.8 : 16.9 & 21.8 : 16.4 & 38.6 : 36.1 & 38.3 : 35.4 & 38.6 : 35.5 & 15.3 : 9.0 & 15.2 : 8.9 & 14.6 : 9.7\\
    $\Delta$ & 4.9 & 5.0 & 5.4 & 2.6 & 2.9 & 3.1 & 6.3 & 6.3 & 5.0\\\hline
    BDCI : KI & 21.1 : 17.1 & 20.6 : 17.3 & 21.2 : 16.5 & 37.2 : 36.3 & 37.5 : 36.2 & 37.6 : 35.5 & 14.3 : 9.8 & 13.9 : 9.7 & 14.2 : 9.4\\
    $\Delta$ & 4.0 & 3.4 & 4.7 & 0.9 & 1.3 & 2.1 & 4.4 & 4.1 & 4.8\\\hline
  \end{tabular}}
  \caption{Tournament results for each game type. Shown are average tournament
    scores per game for players NI (No Inference), BDI (Bidding-Declaration
    Inference), BDCI (Bidding-Declaration-Cardplay Inference), and KI
    (Kermit's Inference) which were obtained by playing 2,500 matches in each
    matchup. Each match consists of two games with soloist/defender roles
    reversed. One standard deviation, averaged over all matchups in a game type,
    amounts to 1.0, 1.4, and 1.0 tournament points per game for null, grand, 
    and suit games respectively.}
  \label{tab:results}
\end{table*}

It is clear that, in terms of the likelihood of sampling the true state, BDCI 
is the strongest of the algorithms considered.
The other two algorithms perform similarly, with BDI having the edge
as soloist, and KI having the edge as defender. 
Across all graphs and players, there are two main phenomena affecting \textit{TSSR}. 
The first is the exponential decrease in the number of states per information set 
and the corresponding decrease of \textit{TSSR}. 
Information sets get smaller as the game progresses and more private information
is revealed --- rapidly driving up the probability of selecting the true 
state with a random guess.
The second is the benefit of using card history for inference, which can be
seen through BDCI's \textit{TSSR} performance compared to the other algorithms
we tested. 
The combination of both of these phenomena is evident in the plots of BDCI, as
the  effect of card history dominates until around trick 6, and then the 
exponential  decrease in the number of states per information set starts to 
equalize the inference capabilities of all techniques. 
 
\subsection{Skat Cardplay Tournaments}

\begin{table}
  \caption{Overall soloist win percentage for each game type across all 
  opponents on test set of games played by humans. 
  Grand games in our test set have a high success rate for the soloist.}
  \label{tab:win_rate}
  \begin{center}
  \begin{tabular}{c|c|c|c}
    Player & Suit & Grand & Null \\
    \hline
    BDCI  & 81.5 & 92.5 & 64.0 \\
    KI    & 80.5 & 92.6 & 62.3 \\
    BDI   & 79.2 & 92.2 & 61.4 \\
    NI    & 78.0 & 91.9 & 59.0 \\
  \end{tabular}
  \end{center}
\end{table}

Tournaments are structured so that pairwise comparisons can be made between
players.
Two players play 2,500 matches in each matchup, and each match consists of two 
games. 
Each player gets a chance to be the soloist against two copies of the other 
player as defenders. 
The games start at the cardplay phase --- with the bidding, discard, and 
declaration previously performed by human players on DOSKV. 
These games are separate from the sets that were used as part of the training 
process, and we calculate results separately for each of Skat's game types.

Table~\ref{tab:results} shows results from each tournament type.
The positive effect of our sampling technique is clearly shown in null and suit 
games, with a statistically significant points increase for BDCI against 
current state-of-the-art KI in these game types.
The lack of statistical significance in grand games for this matchup can be
explained by the high soloist winning percentage for grands in the test set.
Considering that the difference in tournament points per game between KI and
NI is not significant either, it seems that inference cannot overcome the 
overwhelming favorability toward the soloist in our test set of grand games.

NI's overall performance suggests that some type of inference is undoubtedly 
beneficial to PIMC players in suit and null games, but the effect of inference 
in grands is less noticeable in general.
The tournaments between BDCI and BDI suggest that move history helps the 
predictive power of our networks, which in turn causes a substantial increase 
in playing strength.
Results for matches between KI and BDI show that Kermit's
count-based approach may be slightly more robust than using individual card
predictions, but only when the full move history is not considered.

Game type effects are observed in the tournament setting; grand games
have a substantially higher soloist win percentage and null games seem the 
most difficult for our PIMC players.
Conservative human play is the main cause of inference seeming less important
in grands.  In Skat, grand games are worth more than any other game type.
They offer a hefty reward when won, but an even larger penalty when lost.
This explains why we observe that human players are overly-conservative when
it comes to bidding on and declaring grands; they only take a chance and play
when the game is easy to win.
Therefore in the games from our test set, good players won't benefit as much 
from  superior play because the games are too easy for the soloist and too 
hard for the defenders for skill to make a difference. 
This is supported by the overall soloist win percentages for each player shown
in Table~\ref{tab:win_rate}.

A similar explanation can be made for the surprising difficulty all of our 
players seem to have playing null games as the soloist. 
Null games have one of the smallest base values for winning or losing in Skat,
and they have no possibility of additional multipliers.
So when players gamble on the contents of the skat and bid too high relative
to their hand, they will often play null games because they are the cheapest
to lose.
These are notoriously difficult to win because the soloist's hand would
typically contain cards tailored toward the suit or grand game that they bid on, 
whereas a successful null requires completely different cards.

BCDI's superior performance comes with the cost of taking longer to choose
moves than KI.
In matches between BCDI and KI with 320 states evaluated per move, BDCI takes
an average of 0.286 seconds to take an action whereas KI only takes an average
of 0.093 seconds per move --- a 3.1x slowdown.
However, BDCI is still fast enough that it is feasible to play tournaments with
humans in a reasonable amount of time.
Furthermore, KI was shown to reach a performance saturation point after 
sampling 160 states per move \cite{furtak2013recursive},  so increasing the 
time available to KI would not affect the results reported in this work.

\subsection{Discussion}

BDCI's cardplay tournament performance is an exciting result
considering that Kermit (KI) was already judged as comparable to 
expert human players \cite{buro2009improving}.
Additionally, a per-game tournament point increase of more than 4 in suit
and null games means that the gap between the two players is substantial.

Decomposing $p(s|h)$ into a product of individual card location probabilities
(Equation~\ref{eq:world_prob_single}) is a useful approximation. 
First and foremost, it makes it tractable to include entire move histories in 
the context. 
Even when lacking the predictive power of move history (BDI), the method 
still provides some degree of disambiguation between likely and unlikely states. 
However, move history clearly impacts what we can infer about hidden 
information in trick-based card games like Skat.

Knowing where cards lie allows a search-based player to spend more of its
budget on likely states. 
From the tournament results, it is clear that this has a positive effect on 
performance. 
Evaluation is comparatively expensive, so sampling is usually the only option. 
However, even if players could evaluate all states, evaluations would still need to 
weighed by state likelihoods to obtain accurate move values.

Extending beyond what we have shown here, it is our belief that the
effectiveness of our inference technique is not limited to simple evaluation
techniques like PIMC and could be applied in games other than Skat.
IIMC samples states from the root information set before estimating the value
of each move in them by simulating to the end of the game with a playout module.
Applying our technique would result in more realistic states being simulated
by the playout module.
Furthermore, if the playout module is a player that samples and evaluates states
as well, it could also take advantage of our technique to improve the value 
estimates returned to the top-level player.
Applying our technique to ISMCTS is similarly straightforward because the
algorithm samples a state from the root information set before each iteration.
ISMCTS proceeds by only using actions that are compatible with this sampled 
state, so better sampling should cause ISMCTS to perform more realistic 
playouts and achieve more accurate move value estimates.
Adapting our technique to games other than Skat simply requires training a 
neural network with game-specific input features.
As explained in Equation~\ref{eq:world_prob_single}, network output size 
must be defined by the number of cards $|C|$ and the number of 
possible locations for each card $l$ according to the rules of the game.


\section{Conclusions and Future Work} \label{sec:conc}

In this paper we have shown that individual card inference trained by
supervised learning can improve the performance of PIMC-based players in
trick-based card games considerably.
This may not come as a surprise to seasoned Contract Bridge or Skat players as
they routinely draw a lot of information about the whereabouts of remaining
cards from past tricks.
However, this paper demonstrates how to do this using modern learning techniques
for the first time.
It shows how neural networks trained from human data can be used to predict fine-grained
information like the locations of individual cards.
Lastly it shows how to incorporate such predictions into current 
state-of-the-art search techniques for trick-taking card games --- improving an 
already-strong Skat AI system significantly in the process.

This result is exciting and opens the door for further improvements.
Playing strength could  be increased by further improving inference so 
that the model can adjust to individual opponents.
State probability distributions could be smoothed to account for opponents who
often make mistakes or play clever moves to confuse inference.
Furthermore, creating strong recursive IIMC players \cite{furtak2013recursive}
should be possible by incorporating effective inference into the top-level 
player as well as its low-level rollout policies.
This has the potential to overcome some of the key limitations of PIMC players
and potentially achieve superhuman level in trick-taking card games.



\end{document}